\documentclass{bmvc2k}

\usepackage{times}
\usepackage{epsfig}
\usepackage{graphicx}
\usepackage{amsmath}
\usepackage{amssymb}
\usepackage{caption}
\usepackage{multicol}
\usepackage{algpseudocode} 

\usepackage{xcolor}
\usepackage{amssymb}
\usepackage{amsmath}
\usepackage{caption}
\usepackage{booktabs}
\usepackage{makecell}
\usepackage{multirow}
\usepackage{stackengine}
\usepackage{scrextend}
\usepackage[linesnumberedhidden,ruled]{algorithm2e}

\usepackage{blindtext}

\newcommand{\bg}{\ensuremath{{\bold g}}}

\newcommand{\bbf}{\ensuremath{{\bold f}}}

\newcommand{\bbg}{\ensuremath{{\bold g}}}

\newcommand{\mcalG}{\ensuremath{{\mathcal{G}}}}

\newcommand{\mcalX}{\ensuremath{{\mathcal{X}}}}
\newcommand{\mcalC}{\ensuremath{{\mathcal{C}}}}

\newcommand{\mcalD}{\ensuremath{{\mathcal{D}}}}

\DeclareMathOperator*{\argmax}{argmax} 

\SetKwBlock{Repeat}{repeat}{}
\SetKwInOut{Initialization}{Initialization}

\title{Context-Aware Unsupervised  Clustering  \\ for Person Search}

\addauthor{Byeong-Ju Han$^*$}{bjhan@unist.ac.kr}{1}
\addauthor{Kuhyeun Ko$^*$}{khko@unist.ac.kr}{1}
\addauthor{Jae-Young Sim$^\dagger$}{jysim@unist.ac.kr}{1}

\addinstitution{
	Ulsan National Institute of Science and Technology, Ulsan, Republic of Korea 
	\\[4pt]
	$*$ Equal contribution
	\\[4pt]
	$\dagger$ Corresponding author
}

\runninghead{Han, Ko, Sim}{Context-Aware Unsupervised Clustering for Person Search}

\begin{document}
	
	\maketitle
	
	\begin{abstract}
		The existing person search methods use the annotated labels of person identities to train deep networks in a supervised {manner that} requires a huge amount of time and effort for human labeling. In this paper, we first introduce a novel framework of person search that is able to train the network in the absence of the person identity labels, and propose efficient unsupervised clustering methods to substitute the supervision process using annotated person identity labels.  Specifically, we propose a hard negative mining scheme based on the uniqueness property that only a single person has the same identity to a given query person in each image. We also propose a hard positive mining scheme by using the contextual information of co-appearance that neighboring persons in one image tend to appear simultaneously in other images. The experimental results show that the proposed method achieves comparable performance to that of the state-of-the-art supervised person search methods, and furthermore outperforms the extended unsupervised person re-identification methods on the benchmark {person search datasets}.
	\end{abstract}
	
	\vspace{-3mm}
	\section{Introduction} 
	
	Person search has drawn much attention in diverse applications from private entertainment to public safety. However, it is a challenging task to detect person objects from gallery images and to recognize the identities of detected persons together. The existing methods for person search mainly try to train deep networks to embed a distinct feature for each person identity in a supervised manner by using manually annotated identity labels. Human labeling requires much time and effort and often causes incomplete annotation that usually degrades the performance of person search. {A few attempts have} been made to assign existing valid labels to such {unlabeled} persons~\cite{shi2018instance,dai2020dynamic}, {however,} they still follow {the} supervised learning framework and their performance mainly depends on the labeled data.

	\begin{figure}[t]
		\centering
		\includegraphics[width=0.9\textwidth]{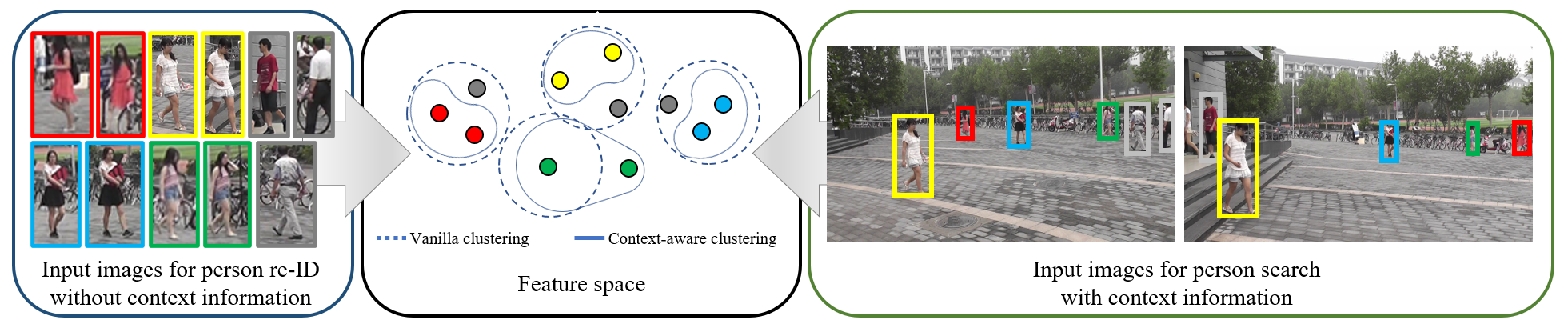}
		\vspace{4mm}
		\caption{{Comparison between the proposed clustering using contextual priors and the vanilla clustering only depending on feature similarity.}}
		\label{fig:intro}\vspace{-4mm}
	\end{figure}

	In this work, we define a novel problem of person search in the absence of labeled person identities and propose an end-to-end network which is trained by context-aware clustering methods to group the persons considered to have the same identity. Specifically, we exploit two specific context properties of \textit{uniqueness} that no more than a single person has the same identity to a given query person in each gallery image and \textit{co-appearance} that multiple persons appeared in an image are highly probable to appear simultaneously in other images. Figure~\ref{fig:intro} conceptually shows the strength of the proposed clustering method using the context properties compared to a vanilla clustering method grouping the samples within a certain feature distance into the same identity. While the vanilla clustering method incorrectly predicts the identities of the persons in the gray boxes due to their similar appearances to others, the proposed clustering method removes them by the uniqueness property. Moreover, whereas the vanilla clustering method fails to group the persons in the green boxes together due to a large feature distance between two green points in the feature space, the proposed method finds such hard positive samples according to the co-appearance property.
	
	The main contributions of this paper are summarized as follows.
	\begin{enumerate}
		\vspace{-1mm}\item We first introduced the problem of person search without identity labels, and proposed context-aware unsupervised clustering methods to solve this problem by using the uniqueness and co-appearance properties of the person search framework. 
		\vspace{-2mm}\item We extended the existing unsupervised person re-ID methods and showed that the proposed method outperforms them and also achieves comparable performance to the state-of-the-art supervised person search methods.
	\end{enumerate}
	
	\vspace{-6mm}
	\section{Related Works}
	This section summarizes the existing works related to the problem of training a person search method in the absence of the person identity labels. We firstly introduce the unsupervised person re-identification methods embedding distinct person features from cropped person images without the person identity labels, and then explain the supervised person search methods especially using context information of scene images.

	\noindent\textbf{Person re-ID for unlabeled data.} The person re-ID methods take cropped images of detected person proposals as input and classify them into clusters according to {the} same person identities. To train person re-ID networks in the absence of identity labels, many unsupervised methods adopted the domain adaptation scheme that aims to make a model trained in a source dataset with identity labels perform well on a target dataset without identity labels~\cite{deng2018image,zhong2019invariance,fu2019self,fan2018unsupervised,wang2018transferable,wei2018person,qi2019novel}. Recently, unsupervised person re-ID methods have been developed without using any prior knowledge associated with {other} datasets. Lin et al.~\cite{lin2019bottom} proposed a bottom-up clustering method that first assigns different clusters to all samples and gradually merges the closest neighboring clusters together into one cluster. However, the merging criterion based on the shortest feature distance {between} two clusters is sensitive to outliers {on cluster boundaries}. Ding et al.~\cite{ding2019dispersion} alleviated this problem by {considering all the samples in each cluster to measure the feature distance between two clusters and Zeng et al.~\cite{zeng2020hierarchical} maximized the shortest feature distance between positive samples and selected hard negative samples.} Wang et al.~\cite{wang2020unsupervised} used a binary vector based multi-label representation to indicate multiple samples with the same person identity. Lin et al.~\cite{lin2020unsupervised} softened the binary values of the multi-label representation according to the relation between samples to avoid hard-decision errors. {While the recent unsupervised re-ID methods mainly adopt advanced distance metrics or soften labeling to reduce errors caused by hidden outliers, the proposed method analyzes the scene context to detect the outliers and create reliable clusters. }

	\noindent\textbf{Person search using image context}  
	{Person search addresses the problem to find persons from a gallery of multi-view scene images, who have the same identity to a query person. All existing methods follow a fully supervised learning framework to detect person proposals in the scene images and classify them into different clusters according to {the guidance of annotated person identity labels.} {Unlike person re-ID, person search can investigate the context information such as uniqueness and co-appearance in the scene images. Dai et al.~\cite{dai2020dynamic} utilized the uniqueness property to temporally assign distinct pseudo labels to each of the unlabeled persons caused by incomplete annotation. Li et al.~\cite{li2021sequential} considered all possible matching pairs between a query person in a query set and a candidate person in gallery images, and found {an optimal set of matching pairs by a bipartite matching algorithm, to preserve the uniqueness property. Yao et al.~\cite{9265450} {especially} proposed a new definition of similarity between persons to enforce the persons in an image not to match the same target person in gallery images. Yan et al.~\cite{Yan_2019_CVPR} constructed a graph whose nodes and edges are defined by detected persons in a pair of images and similarities of the initially matched person pairs, respectively,} to update the matching scores {according to the co-appearance property.} {Compared to the existing methods {that adopt the image context to support the supervised training process or enhance the matching quality on the test,} we {first} apply it on the unsupervised framework and specifically investigate that the properties of uniqueness and co-appearance complementally contribute to training a person search method in the absence of person identity labels.}}}

	\begin{figure*}[!t]
		\centering
		\includegraphics[width=0.85\textwidth]{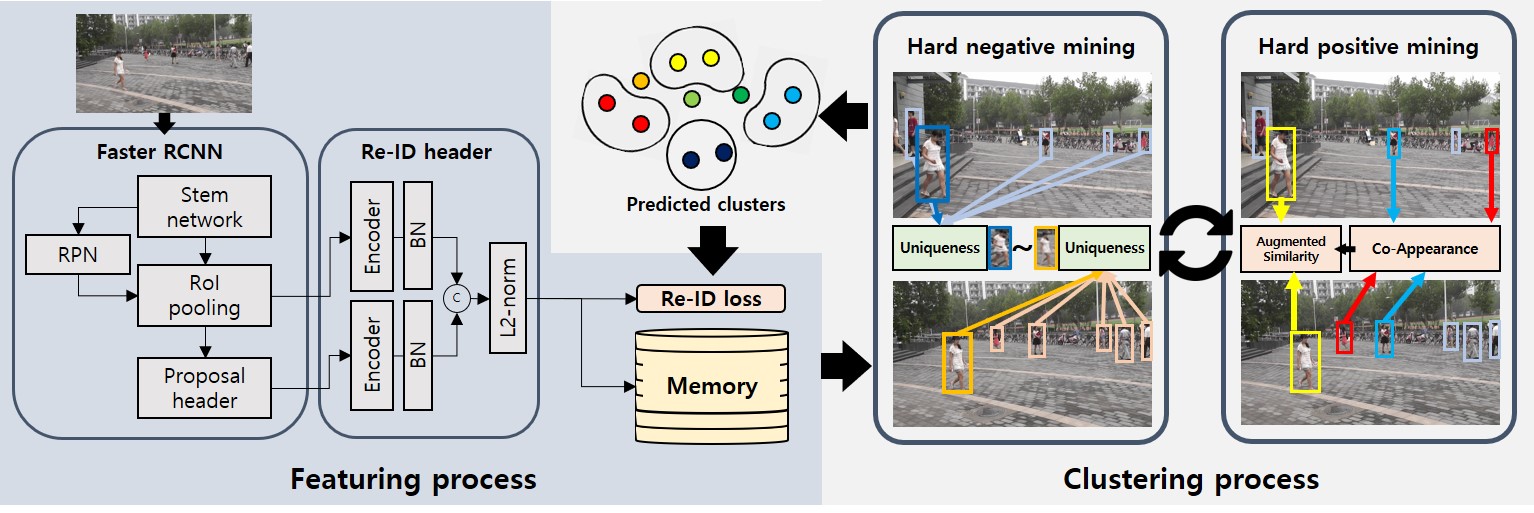}
		\vspace{4mm}
		\caption{Overview of the proposed method. }
		\label{fig:overview}
		\vspace{-6mm}
	\end{figure*}
	
	\vspace{-4mm}
	\section{Proposed Method}\vspace{-0.5mm}
	
	The conventional supervised person search methods train the featuring {process} composed of Faster R-CNN~\cite{faster-rcnn} and {a re-ID header} {using {person} identity labels.} In this work, we introduce a novel problem o{f p}erson search where {the bounding box labels are available for person detection but the person identity labels are not given for person re-ID.} Different from the unsupervised person re-ID problem, we can exploit the contextual information of the persons located within a gallery image to address this problem. {Figure~\ref{fig:overview}} visualizes the overall procedures of the proposed method that contains key clustering modules of hard negative mining (HNM) and hard positive mining (HPM).
	
	\vspace{-3mm}
	\subsection{Uniqueness Based Hard Negative Mining}
	{The existing unsupervised person re-ID methods~\cite{lin2019bottom,lin2020unsupervised} that construct clusters of the same size regardless of person identities may produce limited performances on real datasets because each person identity has diverse numbers of samples as in the person search dataset PRW.} We propose a more reliable clustering method for unsupervised person re-ID, called uniqueness based hard negative mining (HNM), that effectively reduces hard negative samples using the uniqueness property of person search. 
	
	Let $\mcalG = \{ I_1, I_2, \cdots, I_N \}$ be a gallery of images where $N$ is the number of total images, and $\mcalX_l$ denote a set of persons $x_j^l$'s detected in $I_l$. Assume that a query is selected as the $i$-th person $x^k_i$ from the $k$-th image $I_k$, and we find a positive sample of {the query} from the $l$-th image $I_l$ {with $l$ $\neq$ $k$}. We first take the candidate persons in $I_l$, whose feature similarities to the query person are higher than a certain threshold $\delta$, and initialize the set of positive samples as 
	\begin{equation}
		\hat{\mcalC}_l(x^k_i) = \{ x^l_j \mid {s(\bbf^k_i,\bbf^l_j)} > \delta, ~~x^l_j \in {\mcalX_l} \},
	\end{equation}
	where $\bbf^l_j$ is the feature vector of $x^l_j$ stored in the feature memory and $s(\bbf,\bg)$ is the similarity between two features of $\bbf$ and $\bg$. We may collect $\hat{\mcalC}_l(x^k_i)$'s over all the gallery images and take the union $\hat{\mcalC}(x^k_i) = \bigcup_l \hat{\mcalC}_l(x^k_i)$ as a set of positive samples for {the} query $x^k_i$. However, this simple thresholding scheme often includes negative samples to the clusters when relatively high similarity values are computed between two persons of different identities. 
	
	In order to reject such hard negative samples from the clusters, we exploit the uniqueness property of person search that there are no more than a single person in each image having the same identity to a given query person. In practice, we implement this constraint by winner-take-all (WTA) scheme such that only the corresponding sample 
	\begin{equation}
		x^l_{j^*} = \argmax_{x^l_j \in \hat{\mcalC}_l(x^k_i)} {s(\bbf^k_i, \bbf^l_j)},
		\label{eq:forward_max}
	\end{equation}
	which has the highest similarity to the query $x^k_i$, is remained in $\hat{\mcalC}_l(x^k_i)$ and the other samples are removed from $\hat{\mcalC}_l(x^k_i)$. We then have an updated candidate set as {$\mcalC_l(x^k_i) = \{ x^l_{j^*} \}$}. 
	
	\begin{figure}[t]
		\centering
		
		\includegraphics[width=0.9\textwidth]{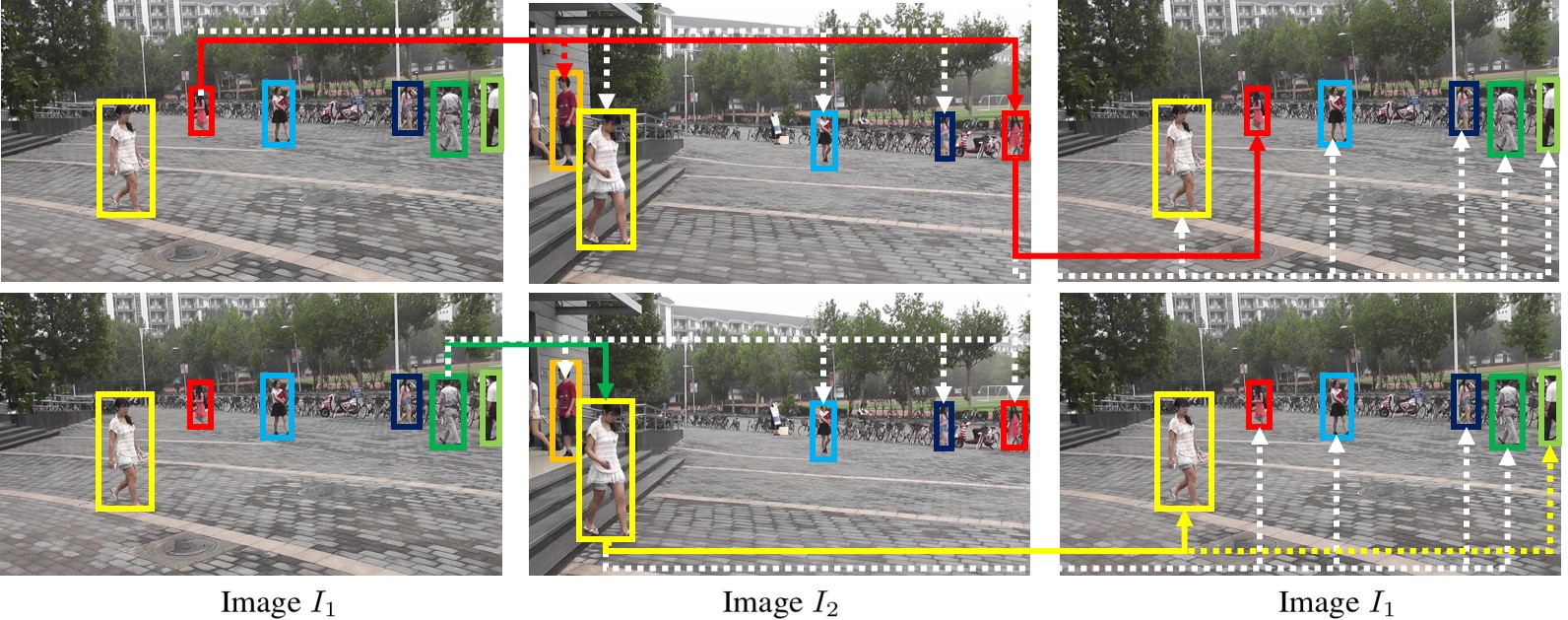}
		\vspace{4mm}
		\caption{{Behaviors} of the uniqueness based HNM. The white and colored lines represent low and high feature similarities between two persons, respectively. The solid lines represent the pairs of the highest feature similarity.}
		\label{fig:consistency}
		\vspace{-4mm}
	\end{figure}
	
	To further improve the accuracy of clustering, we additionally apply the backward matching of WTA scheme from $I_l$ to $I_k$ to check whether the obtained candidate {$x^l_{j^*}$} satisfies the cycle consistency to $x^k_{i}$ or not. If the original query $x^k_i$ becomes the element of $\mcalC_k(x^l_{j^*})$, then we decide that the pair of $x^k_i$ and $x^l_{j^*}$ {has} the same identity to each other and include them into {the} same cluster. Otherwise, we eliminate {$x^l_{j^*}$} from $\mcalC_l(x^k_i)$ due to the violation of the cyclic consistency condition. We decide that there exists no corresponding person to $x^k_i$ in $I^l$ when $\mcalC_l(x^k_i)$ is empty. Finally, we have {a  positive set of the query} $x^k_i$ as {$\mcalC(x^k_i) = \bigcup_l \mcalC_l(x^k_i)$}.
	
	Figure~\ref{fig:consistency} shows the behavior of the proposed uniqueness based HNM where the persons are localized by the bounding boxes in different colors according to their identities. {As shown in the first row of Figure~\ref{fig:consistency},} the person in the red box in {$I_1$} is selected as a query. The two persons in the orange and red boxes are initially detected from {$I_2$} as candidate positive {samples} since they wear red clothes and exhibit higher feature similarities to the query than a threshold. But we select the correct person in the red box in {$I_2$} by WTA which has the highest similarity to the query. The matched pair is finally considered as a positive pair since the original query person in $I_1$ is selected to have the highest similarity to the person in the red box in $I_2$ via the backward matching. On the other hand, as shown in {the second row of Figure~\ref{fig:consistency}}, when selecting the person in the green box in {$I_1$} as a query, the person in the yellow box is detected from {$I_2$} by~\eqref{eq:forward_max} despite not being a true positive sample since the query person does not appear in $I_2$. The person in the yellow box is detected from {$I_1$} via the backward matching instead of the original query person in the green box, and therefore we remove this hard negative sample from the clusters. {Note that} the uniqueness based HNM cannot be applied to the unsupervised person re-ID framework due to the lack of contextual information of the associated gallery images, but can be used in the person search framework. 	
	
	\begin{figure}[t]
		\centering
		\subfigure[PRW]{\includegraphics[height=0.16\textheight]{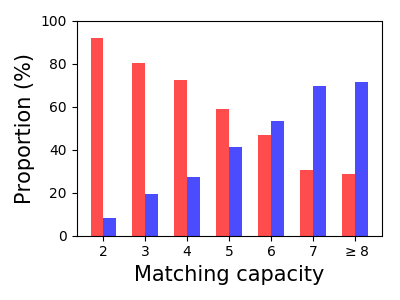}}		
		\hspace{7mm}
		\subfigure[CUHK-SYSU]{\includegraphics[height=0.16\textheight]{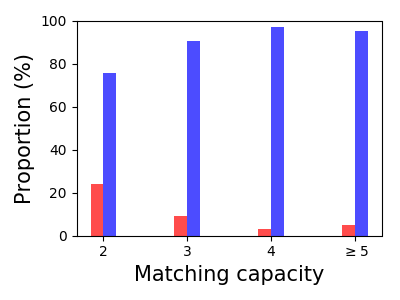}}
		\vspace{2mm}		
		\caption{{Proportions of the image pairs containing a single true positive pair of persons (red) and multiple true positive pairs of persons (blue) according to the matching capacity. (a) PRW and (b) CUHK-SYSU datasets.}}		
		\vspace{-5mm}
		\label{fig:coapp}
	\end{figure}
	
	\subsection{Co-Appearance Based Hard Positive Mining}
	{The uniqueness based HNM for unsupervised clustering still suffers from missing some true positive samples. We also propose a hard positive mining (HPM) scheme that finds challenging positive samples having low feature similarities to the query. We utilize the contextual information of the neighboring persons to the query appeared in {the} same image based on the co-appearance property that multiple persons in an image are likely to appear simultaneously in other images. We investigate this property on PRW and CUHK-SYSU datasets in Figure~\ref{fig:coapp}. We collect all the image pairs containing at least one true positive pair of persons, and classify them into different groups according to the matching capacity that is the minimum number of persons in an image between two images of each pair. Then, for each group, we compare the proportions of the image pairs containing only a single true positive pair of persons where the co-appearance property is not satisfied (red bar) and multiple true positive pairs of persons where the co-appearance property is satisfied (blue bar), respectively. The statistical results in Figure~\ref{fig:coapp} {show} that the number of multiple true positive pairs is increased as more persons appear in an image, and especially CUHK-SYSU dataset exhibits strong co-appearance property.}

	We compute a co-appearance {$A^{(t)}(k,l)$} between $I_k$ and $I_l$ at the $t$-th iteration by considering the clustering results of the neighboring persons given by
	\begin{equation}
		{A^{(t)}(k,l) =  \sum_{x^k_i \in \mcalX_k,~ x^l_j \in \mcalC^{(t)}_l(x^k_i)} s({\bbf^k_i},\bbf^l_j),}
	\end{equation}
	where $\mcalC^{(t)}_l(x)$ is the cluster of $x$ at the $t$-th iteration {on the $l$-th image.}
	Note that {$A^{(t)}(k,l) $} {is obtained by the latest clusters of $\mcalC^{(t)}_l(x)$ and increased} as more persons in $I_k$ are matched to the persons in {$I_l$} with larger similarities. Then, for all the pairs of $x^k_i\in\mcalX_k$ and $x^l_j\in\mcalX_l$, we update the original feature similarity of $s(\bbf^k_i, \bbf^l_j)$ to $s^{(t+1)}(\bbf^k_i, \bbf^l_j)$ at the next iteration as
	\begin{equation} \label{eq:update}
		{s^{(t+1)}(\bbf^k_i, \bbf^l_j) = {s(\bbf^k_i, \bbf^l_j)}  + \beta A^{(t)}(k,l) ,}
	\end{equation} 
	where $\beta $ denotes a weight to adjust the contribution of the co-appearance which is empirically set to $0.1$.  {We also} apply the uniqueness based HNM between $I_k$ and $I_l$ again using the updated feature {similarities} to refine the clustering results. Note that non-empty clusters are maintained since the feature {similarities} are increased by {the} same amount. However, we can check if empty clusters include new elements by detecting hard positive samples using the increased {feature similarities}. In this paper, HPM iterates three times maximally.
	
	\begin{figure}[t]
		\centering
		\includegraphics[width=0.8\textwidth]{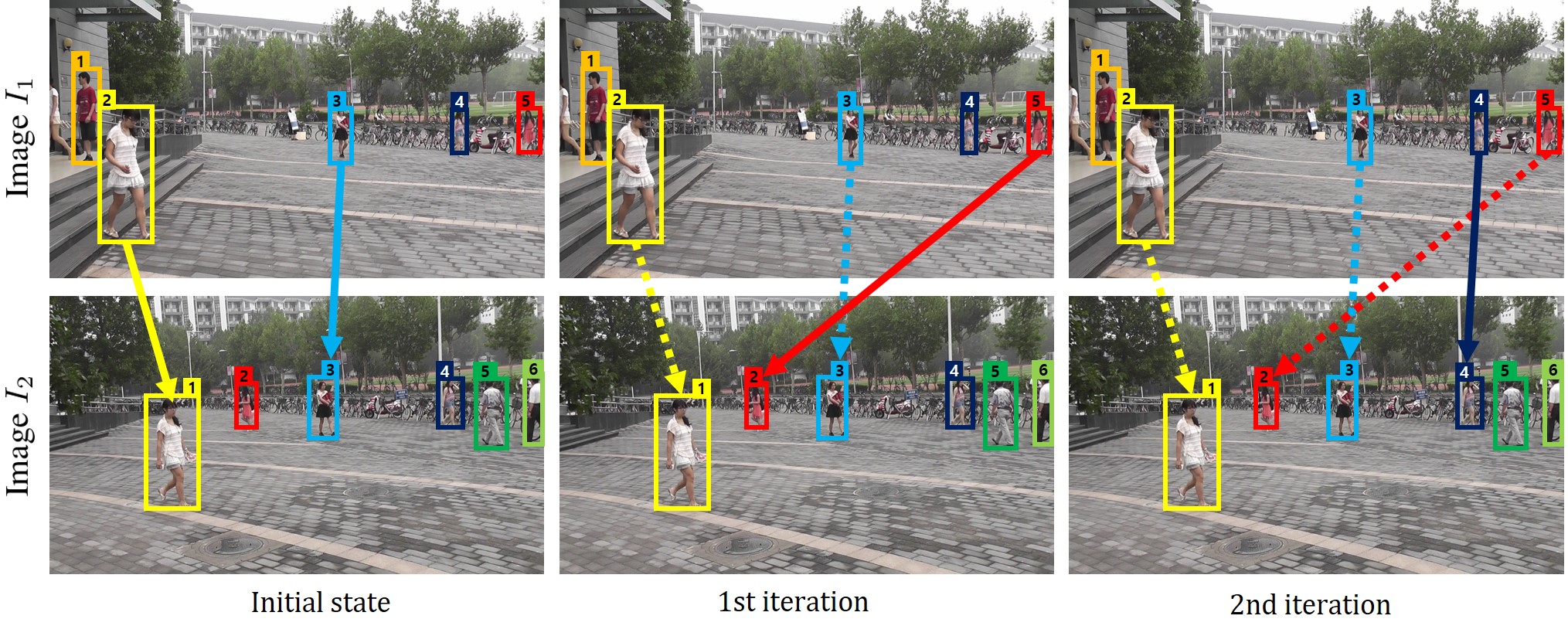}
		\vspace{2mm}
		\caption{{Behaviors} of the co-appearance based HPM. The dotted {arrows} connect the positive pairs detected at the previous states and the solid {arrows} represent the positive pairs newly detected at the current iteration.} 
		\label{fig:coapp_vis}
		\vspace{-5mm}
	\end{figure}
	
	Figure~\ref{fig:coapp_vis} shows two images {$I_1$} and {$I_2$} where the co-appearance based HPM is applied iteratively. The person labels are annotated on the top of the bounding boxes. Let us assume that, at the initial state, the two true positive pairs of {$(x^1_2, x^2_1)$} and {$(x^1_3, x^2_3)$} are correctly clustered, respectively, whereas the other two true positive pairs of {$(x^1_5, x^2_2)$} and {$(x^1_4, x^2_4)$} are missed to be correctly clustered due to relatively low feature similarities. At the first iteration of HPM, {$(x^1_4, x^2_4)$} is still missed to be clustered as a positive pair, but {$(x^1_5, x^2_2)$} is newly detected as a positive pair since the feature similarity $s^{(1)}(\bbf^1_5, \bbf^2_2)$ between $x^1_5$ and $x^2_2$ becomes higher than the threshold $\delta$ by {the co-appearance $A^{(0)}(1,2)$} associated with the persons of {$x^1_2$ and $x^1_3$} neighboring to $x^1_5$. At the second iteration, the additional positive pair {$(x^1_5, x^2_2)$} further increases the co-appearance {to} $A^{(1)}(1,2)$, and therefore increases the feature similarity $s^{(2)}(\bbf^1_4, \bbf^2_4)$ accordingly such that the new pair of {$(x^1_4, x^2_4)$} is detected as a positive pair. 
	
	\vspace{-2mm}
	\subsection{Loss Function}
	{The global} feature memory contains the normalized features for all the person instances in gallery images. Each feature vector  $\bbf_i^k\in\mathbb{R}^{256}$ for $x_i^k\in\mcalX_k$ is initialized as a zero vector and then updated {after back-propagating gradients through the global feature memory} by
	\begin{equation}
		{\bbf^k_i \leftarrow \frac{1}{Z} ( \bbf^k_i +\bbg^k_i ),}
	\end{equation}
	where {$\bbg^k_i\in\mathbb{R}^{256}$} denotes the feature vector extracted from a detected bounding box $b^k_i$ spatially overlapping with the ground truth {bounding} box of $x^k_i$ in $I_k$, and $Z$ is the normalization factor such that {$\Vert \bbf_i^k \Vert _2 = 1$}. 
	
	The total loss function is composed of a detection loss and a re-ID loss, where we adopt a detection loss used in Faster R-CNN~\cite{faster-rcnn}. We propose a re-ID loss for a detected bounding box $b^k_i$ given by
	\begin{equation}\label{eq:trainloss}
		\mathcal{L}(b_i^k) = -\frac{1}{|\mcalC(x_i^k)|}\sum_{x_j^l\in\mcalC(x_i^k)} \log\left( p(x_j^l|b_i^k)  \right),
	\end{equation}
	that encourages $b^k_i$ to have the same identity to $x_j^l\in\mcalC(x_i^k)$  by maximizing the probability 
	\begin{equation}
		\!\!\!p(x_j^l|b_i^k) =\frac{\exp\left(s(\bbf_j^l,\bbg_i^k) / \tau \right) }{\sum_{x \in \{\mcalC(x_i^k) \bigcup \mcalD(x_i^k) \}} \exp\left(s(\bbf,\bbg_i^k) / \tau \right)},
		\label{eq:prob}
	\end{equation}	
	where $\mcalD(x_i^k)$ is the set of the negative samples in $\mcalC^c(x_i^k)$ having top 1$\%$ similarities to $x^k_i$, and $\tau$ is a temperature coefficient empirically set to 0.1. Note that the denominator in~\eqref{eq:prob} does not consider the negative samples having relatively low similarities to $x_i^k$, but mainly includes challenging negative samples having high similarities to $x_i^k$. Therefore, the proposed re-ID loss encourages {increasing} the probability associated with the positive samples while effectively decrease the probability for the challenging negative samples. Moreover, the probability $p(x_j^l|b_i^k)$ is equally maximized by $\frac{1}{|\mcalC(x_i^k)|}$ for all $x_j^l\in\mcalC(x_i^k)$ clustered to be {positive} samples of $x^k_i$, and eventually achieves good performance since HNM and HPM cooperate together to include most of the positive samples to $\mcalC(x_i^k)$ successfully while excluding even challenging negative samples from $\mcalC(x_i^k)$ reliably. 
	
	\vspace{-3mm}
	\section{Experimental Results} \label{sec:experimental_results}
	\vspace{-1mm}
	\subsection{Experimental Setup}
	
	\noindent \textbf{Benchmark datasets.} {{CUHK-SYSU~\cite{Xiao_2017_CVPR} and PRW~\cite{zheng2017person} have been widely used as benchmark datasets for person search. CUHK-SYSU dataset is composed of 11,206 images for training and 6,978 images for testing and provides 96,143 bounding boxes with 8,432 identities to indicate individual persons. It also provides 2900 test queries and predetermined galleries of test images. PRW dataset consists of a training set of 5,704 images and a test set of 6,112 images captured at 6 different fixed camera positions. It provides 43,110 bounding boxes of persons with 932 identities including 2,057 test queries.} However, since PRW does not report a detailed description to define galleries to search persons matching to test queries, we define two different types of {the} gallery: \textit{regular gallery} and \textit{multi-view gallery} in this paper.} The regular gallery consists of all the test images except the image where the query is detected, and ignores the unlabeled persons. This is expected to be a similar setup to the gallery used in many existing methods. The multi-view gallery is quite challenging {because it} excludes the redundant test images captured from {the} same camera view to that of the query image and includes all the unlabeled persons. Note that although both datasets provide the identity labels, we do not use them for training the network to assume the unsupervised framework with the absence of identity labels. 
	
	\noindent\textbf{Implementation details.} We use the sequence of blocks from `conv1' to `conv4' of ResNet-50~\cite{resnet50} pre-trained on ImageNet~\cite{imagenet} for image classification as a stem network, and use `conv5' block of ResNet-50 as a proposal header, respectively, in Figure~\ref{fig:overview}. Each encoder of the re-ID header is defined as a fully connected layer to squeeze the high dimensional feature brought from the backbone network to a feature vector of size {128}. {We train the proposed network using the same hyper-parameters for CUHK-SYSU and PRW datasets except for the number of training epoch and the learning rate decay. The training strategy is empirically determined based on the ablation study and related primary research. {More details are described in the supplementary material.}}
	
	\noindent \textbf{Evaluation metrics.} The performance of person search is usually evaluated by the two metrics: mean of Average Precision (mAP) and Top-$k$ score. mAP considers {both} the precision and recall of the predicted results by computing the average area under the precision-recall curves. Top-$k$ score checks whether the predicted top-$k$ candidates best matching to a given query include at least one true positive sample or not.

	\begin{figure}
		\begin{minipage}[b]{0.5\linewidth}
			\centering
			
			\subfigure[]{\includegraphics[height=0.075\textheight]{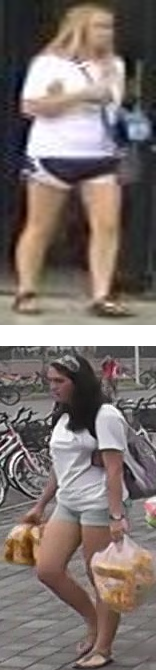}}\hspace{0.5mm}
			\subfigure[]{\includegraphics[height=0.075\textheight]{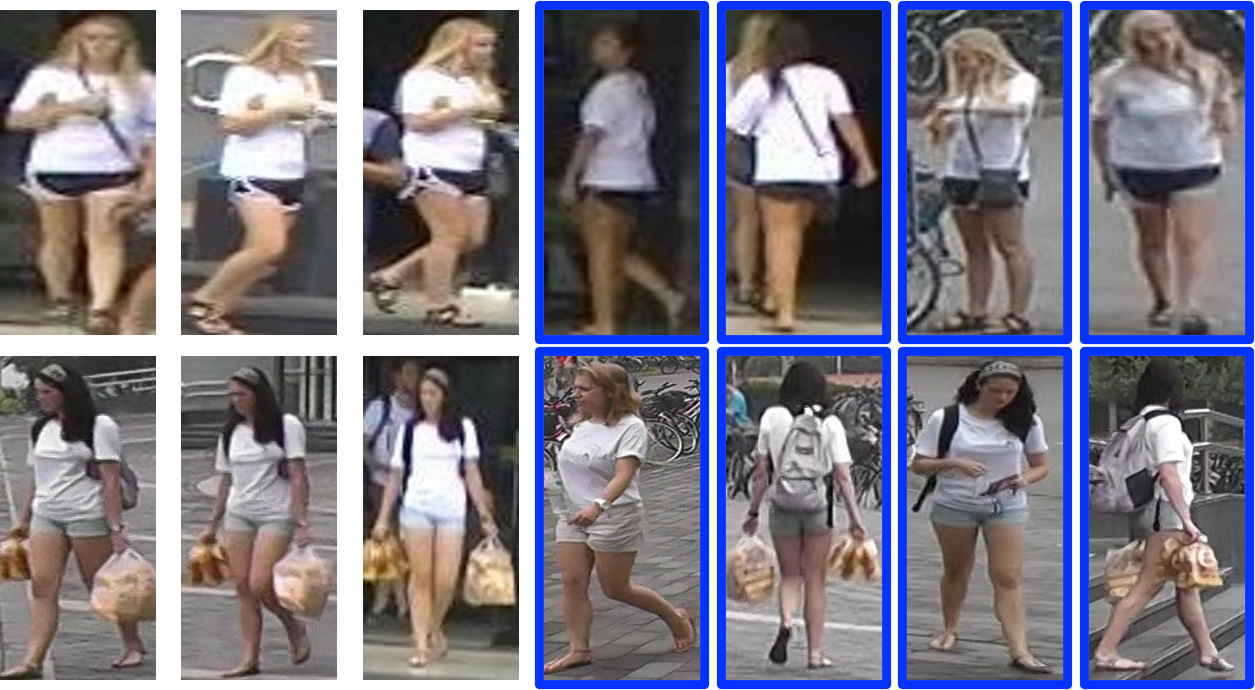}}\hspace{0.5mm}
			\subfigure[]{\includegraphics[height=0.075\textheight]{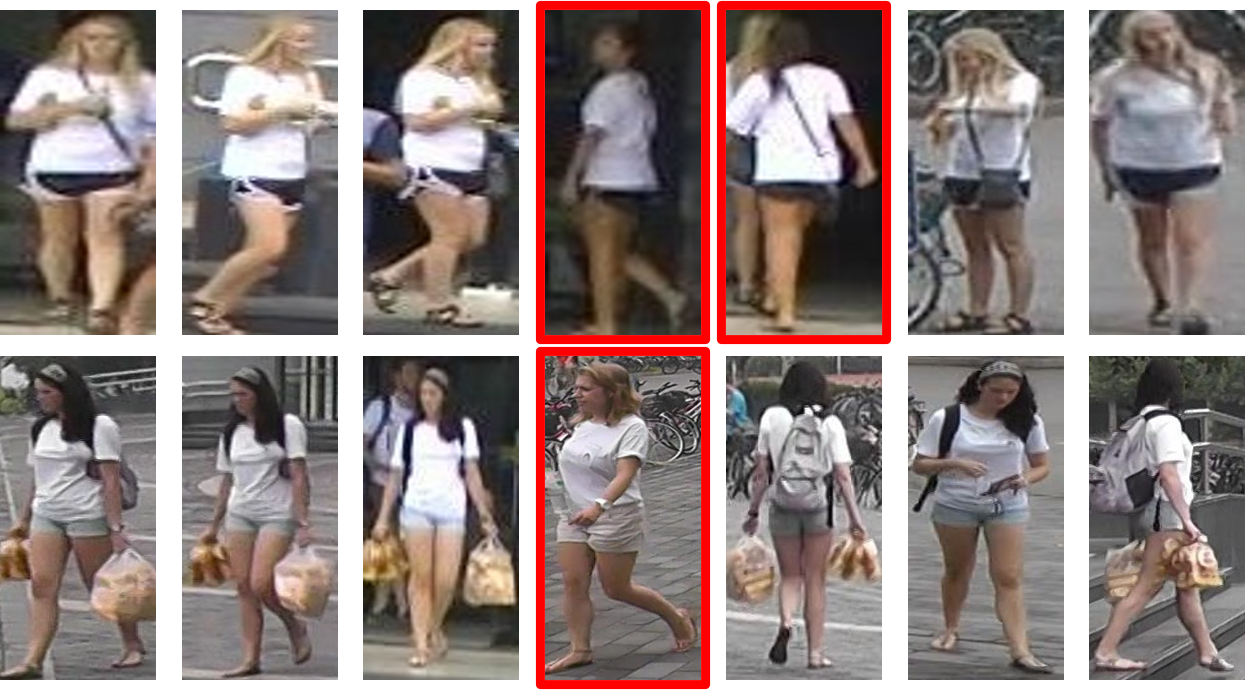}}\vspace{1mm}
			\captionof{figure}{{{Clustering results. (a) Query persons. The persons clustered by using (b) HPM and (c) HPM+HNM.}}}
			\label{fig:clustering_process}
		\end{minipage}\hspace{1mm}
		\begin{minipage}[b]{0.48\linewidth}
			\vspace{-2mm}
			\resizebox{0.99\textwidth}{!}{			
				\begin{tabular}{c|cc|c} 
					\toprule
					& w/o HNM        & with HNM        & HNM gain  \\ 
					\midrule
					w/o HPM        & {27.68/59.35} & {28.01/60.28} & {+0.33/+0.93}  \\
					with HPM       & {32.87/62.86} & {36.61/64.85} & {+3.74/+1.99}  \\ \midrule 
					HPM gain	   & {+5.19/+3.51} & {+8.60/+4.57} & {+8.93/+5.50}  \\
					\bottomrule
				\end{tabular}
			} \vspace{4mm}
			\captionof{table}{Ablation study of the effect of HNM and HPM. Each cell shows the performance in terms of mAP and Top-1 score, respectively.}
			\label{tab:ablation}
		\end{minipage}
		\vspace{-5mm}
	\end{figure}

	\vspace{-3mm}
	\subsection{Ablation Study}
	We conduct {ablation studies} about the performance of the proposed method using the multi-view gallery in PRW dataset.
	
	\noindent \textbf{Co-appearance based HPM.} The proposed HPM is designed to detect challenging positive samples whose feature similarities to a given query are relatively low due to the variation of human poses and/or the changes of camera viewing directions. Each row in Figure~\ref{fig:clustering_process} (b) shows the matched persons to a query person in Figure~\ref{fig:clustering_process} (a) clustered by HPM. We see that HPM adds more persons in blue into the initial cluster of the first three persons by exploiting the co-appearance property. Also, Table~\ref{tab:ablation} shows that the quantitative performance gains of using HPM over the naive method, that generates clusters using a constant threshold for feature similarity, are {5.19} and {3.51} in terms of mAP and Top-1 score, respectively. 
	
	\noindent \textbf{Uniqueness based HNM.} {HNM reduces the false positive errors for clustering by detecting challenging negative samples. Figure~\ref{fig:clustering_process} (c) shows the clustering results of using HPM and HNM together, where we see that HNM removes the unreliable persons in red from the augmented clusters shown in Figure~\ref{fig:clustering_process} (b). As shown in Table~\ref{tab:ablation}, HNM also improves the performance by {3.74} and {1.99} in terms of mAP and Top-1 score, respectively, when used with HPM together, {which means HNM can effectively reduce false positive errors caused by HPM.} Consequently, we see that each of the proposed HNM and HPM clearly demonstrates the effectiveness and achieves a remarkable synergy thanks to their complementary behaviors.} 	
	
	\begin{figure}[t]
		\centering
		\subfigure[]{\includegraphics[width=0.22\textwidth]{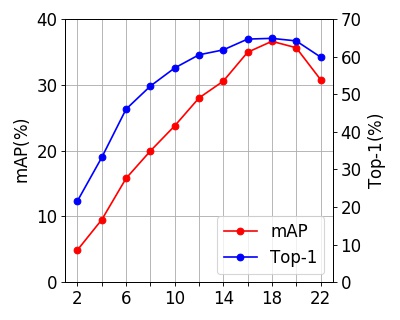}}
		\subfigure[]{\includegraphics[width=0.22\textwidth]{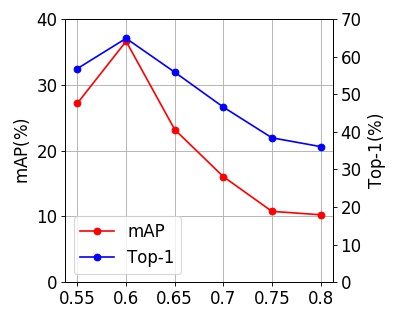}}
		\subfigure[]{\includegraphics[width=0.22\textwidth]{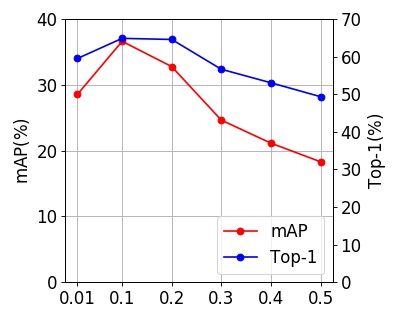}}
		\subfigure[]{\includegraphics[width=0.22\textwidth]{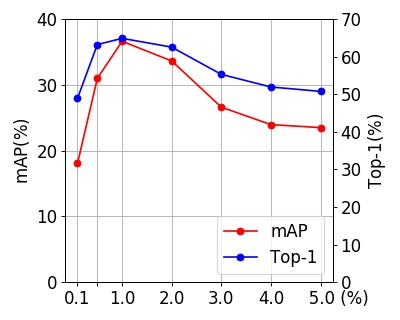}}
		\vspace{2mm}
		\caption{Effects of the hyper-parameters. The performance of the proposed method according to varying (a) the number of training iterations, (b) the threshold for feature similarity, (c) the weight $\beta$ for co-appearance based update of feature similarity, and (d) the ratio of hard negative samples for training loss computation.}
		\label{fig:hyperparameter}
		\vspace{-5mm}
		
	\end{figure}
	
	\noindent\textbf{Number of training iterations.}  Figure~\ref{fig:hyperparameter} (a) shows the performance scores of the proposed method evaluated up to {the 22nd} epoch of training. We see that the performance gradually increases according to the training iteration and reaches the top score at the 18th epoch, but decreases with more epochs due to the over-fitting. 
	
	\noindent\textbf{Threshold for feature similarity.} The HNM module first collects candidate positive samples having higher feature similarities to a given query than a certain threshold. Finding an optimal threshold is important since inappropriately determined boundaries of clusters especially in {the} early training phase may cause incorrect training and eventually result in worse performance. Figure~\ref{fig:hyperparameter} (b) shows the performance {scores varying} this threshold from 0.55 to 0.8, where we see that the best performance is achieved with the threshold value of 0.6. 
	
	\noindent\textbf{Weight $\beta$.} Figure~\ref{fig:hyperparameter} (c) compares the performance of the proposed method according to different weights of $\beta$ in~\eqref{eq:update} used for {the} iterative update of the feature similarity. We have the best performance using $\beta=0.1$. Note that HPM with $\beta>0.1$ clusters lots of false positive samples incorrectly that are hard to be detected by HNM. {In contrast}, HPM with $\beta<0.1$ does not sufficiently exploit {the co-appearance property} to detect hard positive samples. 
	
	\noindent\textbf{Ratio of hard negative samples.} Figure~\ref{fig:hyperparameter} (d) visualizes the impact of the ratio of hard negative samples to construct $\mcalD(x_i^k)$ in~\eqref{eq:prob} for training loss computation. The best performance is achieved when we {set the ratio to 1$\%$}. If we select more negative samples, relatively easy negative samples may have more contribution to train the network yielding decreased performance. {In contrast}, if we take {fewer} negative samples, the network may not be properly trained to encourage large feature {distances} between the query and hard negative samples {due to the lack of diversity of negative samples.} 	\vspace{-1mm}

	\subsection{Comparison {with} Existing Methods} \vspace{-1mm}
	
	{Since there are no existing person search methods to handle the absence of person identity labels, the proposed method is compared with the existing supervised person search methods as well as  the extensions of the existing unsupervised person re-ID methods {that are trainable without person identity labels and any prior knowledge}, respectively. For extension, an individually trained person detection network of Faster R-CNN~\cite{faster-rcnn} is followed by each re-ID network. Table~\ref{tab:results} quantitatively compares the proposed method with the six state-of-the-art person search methods~\cite{9003518, han2021decoupled, li2021sequential, Yan_2021_CVPR, Kim_2021_CVPR, 9265450} in CUHK-SYSU and the regular gallery of PRW. And Table~\ref{tab:results2} shows a more detailed performance comparison between the proposed method and the two extended unsupervised state-of-the-art re-ID methods~\cite{lin2019bottom, wang2020unsupervised}, evaluated in CUHK-SYSU and both galleries of PRW.}
	
\begin{figure}[t]
		\centering
		
		\includegraphics[height=0.07\textheight]{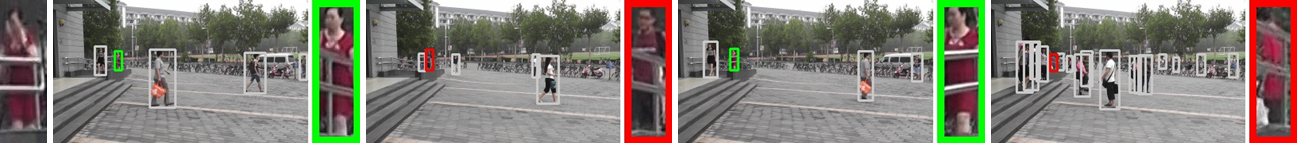}
		\vspace{3mm}
		\caption{{Challenging cases for the proposed unsupervised clustering method. The leftmost figure shows a query person, and the searched persons are highlighted in green and red corresponding to the true and false matching, respectively. }}
		\label{fig:failure_cases}
		\vspace{-5mm}     
	\end{figure}

	\begin{table}[t]
		\begin{multicols}{2}
			\begin{minipage}{0.49\textwidth}
				\centering
				\resizebox{\textwidth}{!}{
					\begin{tabular}{c|c|cc|cc}
						\toprule
						\multirow{2}{*}{Method} & \multirow{2}{*}{Supervised}  & \multicolumn{2}{c|}{CUHK-SYSU} & \multicolumn{2}{c}{PRW} \\  \cline{3-6}
						&     & mAP     & Top-1   & mAP   & Top-1       \\ \midrule
						MGTS~\cite{9003518}              & Yes & 83.3    & 84.2    & 32.8  & 72.1  \\  
						DMRNet~\cite{han2021decoupled}      & Yes & 93.2    & 94.2    & 46.9  & 83.3  \\  
						SeqNet~\cite{li2021sequential}              & Yes & 94.8    & 95.7    & 47.6  & 87.6  \\  
						AlignPS~\cite{Yan_2021_CVPR}              & Yes & 94.0    & 94.5    & 46.1  & 82.1  \\  
						PGA~\cite{Kim_2021_CVPR}              & Yes & 92.3    & 94.7    & 44.2  & 85.2  \\  
						OR~\cite{9265450}              & Yes & 93.2    & 93.8    & 52.3  & 71.5  \\  
						{Proposed}				    & No & {\textbf{81.1}}   & {\textbf{83.2}}    &{\textbf{41.7}}    & {\textbf{86.0}}  \\ 
						
						\bottomrule  
					\end{tabular}
				}
				\vspace{3mm}
				\caption{{Quantitative comparison with the existing supervised person search methods on CUHK-SYSU and regular gallery of PRW.} }
				\label{tab:results}
			\end{minipage}
			
			\begin{minipage}{0.49\textwidth}
				\centering
				\resizebox{\textwidth}{!}{
					\begin{tabular}{c|c|c|c|c|c}
						\toprule
						\multirow{2}{*}{Method}              & \multicolumn{2}{c|}{CUHK-SYSU}    & \multicolumn{3}{c}{PRW}                                        \\ \cline{2-6}
						& mAP             & Top-1           & mAP             & Top-1           &      Gallery               \\
						\midrule
						DET+BUC~\cite{lin2019bottom}         & 74.8            & 77.4            & 26.0            & 83.6            & \multirow{3}{*}{\rotatebox{0}{\footnotesize{Regular}}}    \\
						DET+MLC~\cite{wang2020unsupervised}  & 69.2            & 73.7            & 25.4            & 84.7            &                             \\
						\textbf{Proposed}                    & {\textbf{81.1}} & {\textbf{83.2}} & {\textbf{41.7}} & {\textbf{86.0}} &                             \\
						\midrule 
						DET+BUC~\cite{lin2019bottom}        & -               & -               & 18.6            & 53.0            & \multirow{3}{*}{\rotatebox{0}{\footnotesize{\shortstack{Multi\\-view}}}} \\
						DET+MLC~\cite{wang2020unsupervised} & -               & -               & 17.1            & 50.8            &                             \\
						\textbf{Proposed}                   & -               & -               & {\textbf{36.6}} & {\textbf{64.9}} &                            \\
						\bottomrule
						
					\end{tabular}
				}
				\vspace{3mm}
				\caption{{Quantitative comparison with the extended unsupervised person re-ID methods on CUHK-SYSU and regular and multi-view galleries of PRW.} }
				\label{tab:results2}
			\end{minipage}
			
		\end{multicols}
		
		\vspace{-4mm}
	\end{table}
	{Though trained without the labels of person identity, the quantitative performance of the proposed method achieves more than {85}\% of the top scores of the state-of-the-art supervised methods on both datasets in terms of mAP and Top-1, respectively, as shown in Table~\ref{tab:results}. Furthermore, the proposed method yields significantly better performance than that of the extended unsupervised person re-ID methods in both datasets. Particularly, in Table~\ref{tab:results2} where the lower block shows the performances evaluated in the challenging multi-view gallery of PRW, we see that the proposed method providing the best mAP score of {36.6} almost improves the performance twice compared to the {second-best} method with mAP score of 18.6 on the multi-view gallery. {Note that the proposed method and the extensions of BUC and MLC generate person clusters mainly employ the appearances of the detected persons in unsupervised manners, and therefore they often provide non-negligible similarities between different persons exhibiting similar looks,  especially in PRW dataset that contains many different persons wearing similar cloths as shown in Figure~\ref{fig:failure_cases}. In such cases we have high Top-1 scores but relatively low mAP scores.} Due to the space limit, the qualitative results are reported in the supplementary material.}

	\vspace{-2mm}
	\section{Conclusion}\vspace{-2mm}
	We addressed a novel person search problem without using the labels of person identities in this work. We investigated contextual properties of {the} person search framework, and proposed two unsupervised clustering methods of the uniqueness based HNM and the co-appearance based HPM to classify unlabeled person {samples. We} conducted the comparative experiments including ablation {studies} of the proposed method. Experimental results demonstrated that the proposed method provides comparable performance to the existing state-of-the-art supervised person search methods and outperforms the extended unsupervised person re-ID methods. 
	
	\bibliography{egbib}
\end{document}